\documentclass[10pt,twocolumn,letterpaper]{article}

\usepackage[pagenumbers]{cvpr} 
\usepackage{url}
\usepackage{graphicx}
\usepackage{tabularx}       
\usepackage{comment}
\usepackage{makecell}
\usepackage{pifont}
\usepackage{booktabs}
\usepackage{multirow}

\usepackage[misc]{ifsym}
\usepackage{afterpage}
\usepackage{setspace}
\usepackage{breakcites}
\usepackage{seqsplit}

\usepackage{times}
\usepackage{epsfig}
\usepackage{graphicx}
\usepackage{amsmath}
\usepackage{amssymb}
\usepackage{float}
\usepackage[symbol]{footmisc}

\usepackage{bm}
\usepackage[accsupp]{axessibility} 

\newcommand{\myparagraph}[1]{\vspace{0.1em}\noindent\textbf{#1}}

\usepackage[dvipsnames]{xcolor}

\definecolor{cvprblue}{rgb}{0.21,0.49,0.74}
\usepackage[pagebackref,breaklinks,colorlinks,citecolor=cvprblue]{hyperref}

\title{OMG: Towards Open-vocabulary Motion Generation via Mixture of Controllers}

\vspace{-8pt}
\author{
Han Liang$^{1}$\hspace{2.3em}
Jiacheng Bao$^{1}$\hspace{2.3em} 
Ruichi Zhang$^{1}$\hspace{2.3em} 
Sihan Ren$^{1}$\hspace{2.3em}
Yuecheng Xu$^{1}$\vspace{0.3em} \\
Sibei Yang$^{1}$\hspace{2.3em} 
Xin Chen$^{2}$\hspace{2.3em} 
Jingyi Yu$^{1}$\hspace{2.3em} 
Lan Xu$^{1}$ \vspace{0.5em} \\
$^{1}$\textit{ShanghaiTech University} \hspace{1.3em} 
$^{2}$\textit{Tencent PCG}
}

\vspace{-16pt}

\begin{document}
\maketitle

\begin{abstract}

We have recently seen tremendous progress in realistic text-to-motion generation. 
Yet, the existing methods often fail or produce implausible motions with unseen text inputs, which limits the applications.
In this paper, we present OMG, a novel framework, which enables compelling motion generation from zero-shot open-vocabulary text prompts. 
Our key idea is to carefully tailor the pretrain-then-finetune paradigm into the text-to-motion generation.
At the pre-training stage, our model improves the generation ability by learning the rich out-of-domain inherent motion traits.
To this end, we scale up a large unconditional diffusion model up to 1B parameters, so as to utilize the massive unlabeled motion data up to over 20M motion instances. 
At the subsequent fine-tuning stage, we introduce motion ControlNet, which incorporates text prompts as conditioning information, through a trainable copy of the pre-trained model and the proposed novel Mixture-of-Controllers (MoC) block.
MoC block adaptively recognizes various ranges of the sub-motions with a cross-attention mechanism and processes them separately with the text-token-specific experts.
Such a design effectively aligns the CLIP token embeddings of text prompts to various ranges of compact and expressive motion features. 
Extensive experiments demonstrate that our OMG achieves significant improvements over the state-of-the-art methods on zero-shot text-to-motion generation. 
Project page: \tt \small \textbf{\href{https://tr3e.github.io/omg-page}{https://tr3e.github.io/omg-page}}.

\end{abstract}

\section{Introduction}
\label{sec:intro}

Generating vivid motions of human characters is a long-standing task in computer vision, with numerous applications in movies, games, robotics, and VR/AR. The text-to-motion setting aims to democratize motion generation for novices and has recently received substantive attention.

\begin{figure}[tbp] 
	\centering  
	\includegraphics[width=1\linewidth]{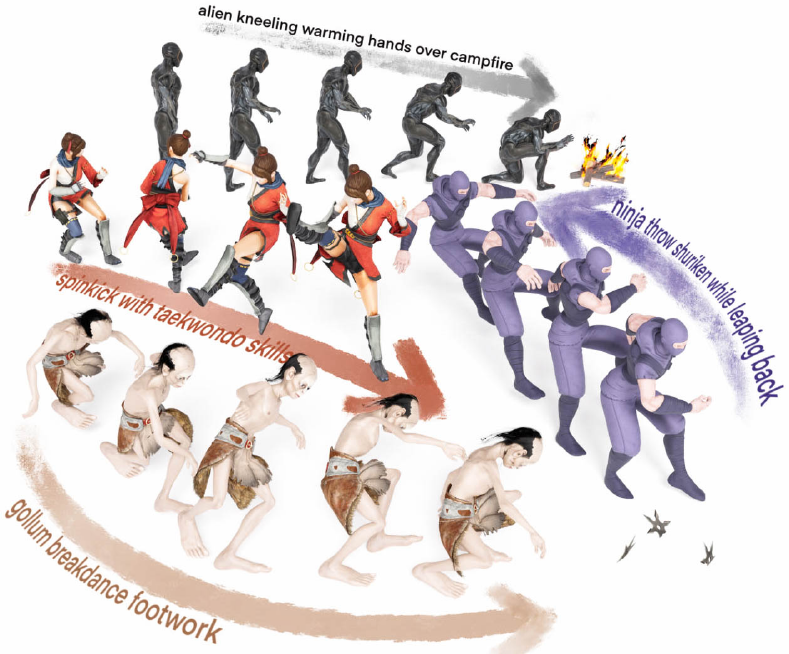} 
	\caption{Our Open-vocabulary Motion Generation (OMG) approach is capable of generating high-quality motions in response to unseen text prompts.}
	\label{fig:teaser} 
\end{figure}

Benefited from the existing text-annotated motion datasets~\cite{Guo_2022_CVPR,mahmood2019amass}, recent advances generate diverse motions from text prompts, equipped with various generative models like VAEs~\cite{petrovich2022temos,tevet2022motionclip}, autoregressive models~\cite{zhang2023t2m,jiang2023motiongpt}, and diffusion models~\cite{tevet2022human,zhang2022motiondiffuse,dabral2023mofusion,chen2023executing}. However, these methods heavily rely on the paired text-motion data with limited text annotations, and hence fall short of generalizing to unseen open-vocabulary text inputs.
Imagine generating ``\textit{Gollum breakdance footwork}'', it requires 
out-of-domain generation ability with rich human knowledge to understand the motion traits in various words like characters (Gollum, ninja, etc.) or skills (breakdance, spinkick, etc.). To tackle the open-vocabulary texts, recent works~\cite{hong2022avatarclip,lin2023being,azadi2023make} utilize the zero-shot text-image alignment ability of the large-scale pre-trained models, e.g., CLIP~\cite{radford2021learning}. They adopt text-pose alignments and hence remove the reliance on pair-wise text and continuous motions. Yet, the discrete poses are insufficient to represent the rich motion traits, leading to unrealistic motion generation. 
On the other hand, recent advances~\cite{rombach2022high,zhang2023adding, guo2023animatediff, videoworldsimulators2024} enable impressive and controllable text-driven image and even video generation, even from open-vocabulary text inputs. Revisiting their huge success, both the pretrain-then-finetune paradigm as well as scaling up the model have played an important role. Yet, adopting these strategies for text-to-motion generation is challenging. First, there exists a huge imbalance between the quantities of unpaired motion data and paired text-motion data. Second, for open-vocabulary texts, various tokens correspond to various motion traits, constituting a complex many-to-many problem.

To tackle the challenges, we propose \textit{OMG} -- a novel scheme to generate motions of human characters from open-vocabulary texts (see \cref{fig:teaser}). Our key idea is to carefully tailor the pretrain-then-finetune paradigm into text-to-motion generation. For pre-training, we adopt a minimalistic design to scale up a large unconditional diffusion model, so as to utilize the massive unlabeled motion data. For fine-tuning, we first freeze the pre-trained large model. We then adopt the trainable copy and treat the text prompts as extra conditions through a novel design, named mixture-of-controllers, to learn to predict the conditional residuals. It adaptively fuses the motion traits corresponding to various tokens from the input text prompts, so as to handle the ambiguity between the text and motion modalities.

Specifically, at the pre-training stage, we adopt the diffusion transformer~\cite{peebles2023scalable} as the backbone with over-parameterized motion representation. We then scale up the models with parameters ranging from 88M to 1B, leveraging over 20M unlabeled motion instances from a diverse range of 13 publicly available datasets. During training, we also adopt a sliding random window strategy to crop various motion sequences, to improve the generation ability for arbitrary lengths of motion. To this end, the pre-trained model learns the rich inherent motion features with a large motion manifold to ensure the realism of the generated motions. For the subsequent stage, to incorporate the text prompts as conditioning information, we adopt a fine-tuning scheme called motion ControlNet, analogous to the famous ControlNet~\cite{zhang2023adding} for the text-image generation task. It includes a trainable copy of the large-scale pre-trained model, which serves as a strong backbone to retain the motion features, as well as a novel block called Mixture-of-Controllers (MoC). Such MoC block can effectively inject residual information into the pre-trained model, based on the motion features and the CLIP-based text embeddings. Our key design in the MoC block is the cross-attention mechanism between text and motion, as well as the text-token-specific Mixture-of-Experts~\cite{shazeer2017outrageously,fedus2022switch}, which are to be detailed in later sections. Such a design effectively aligns the token embeddings of text prompts to various ranges of compact and expressive motion features, which are warmed up from a powerful pre-trained model. 
As a result, our OMG approach achieves compelling generation from open-vocabulary texts, significantly outperforming prior arts. In particular, only fine-tuned on the HumanML3D dataset~\cite{Guo_2022_CVPR}, it can generate vivid motions of various human characters with diverse motion skills from the Mixamo dataset~\cite{Mixamo}, as shown \cref{fig:teaser}.

To summarize, our main contributions include: 
\begin{itemize} 
\setlength\itemsep{0em}
    \item We propose a text-to-motion generation approach with a pretrain-then-finetune paradigm to scale up both data and model, achieving state-of-the-art zero-shot performance. 
   
    \item We experimentally demonstrate that pre-training on large-scale unlabeled motion data improves the generation results from diverse and open-vocabulary texts.
   
    \item We propose a fine-tuning scheme for text conditioning, utilizing a mixture of controllers to effectively improve the alignment between text and motion.

\end{itemize}

\section{Related Work}
\label{sec:related}

\begin{figure*}[htbp] 
	\centering  
	\includegraphics[width=0.985\linewidth]{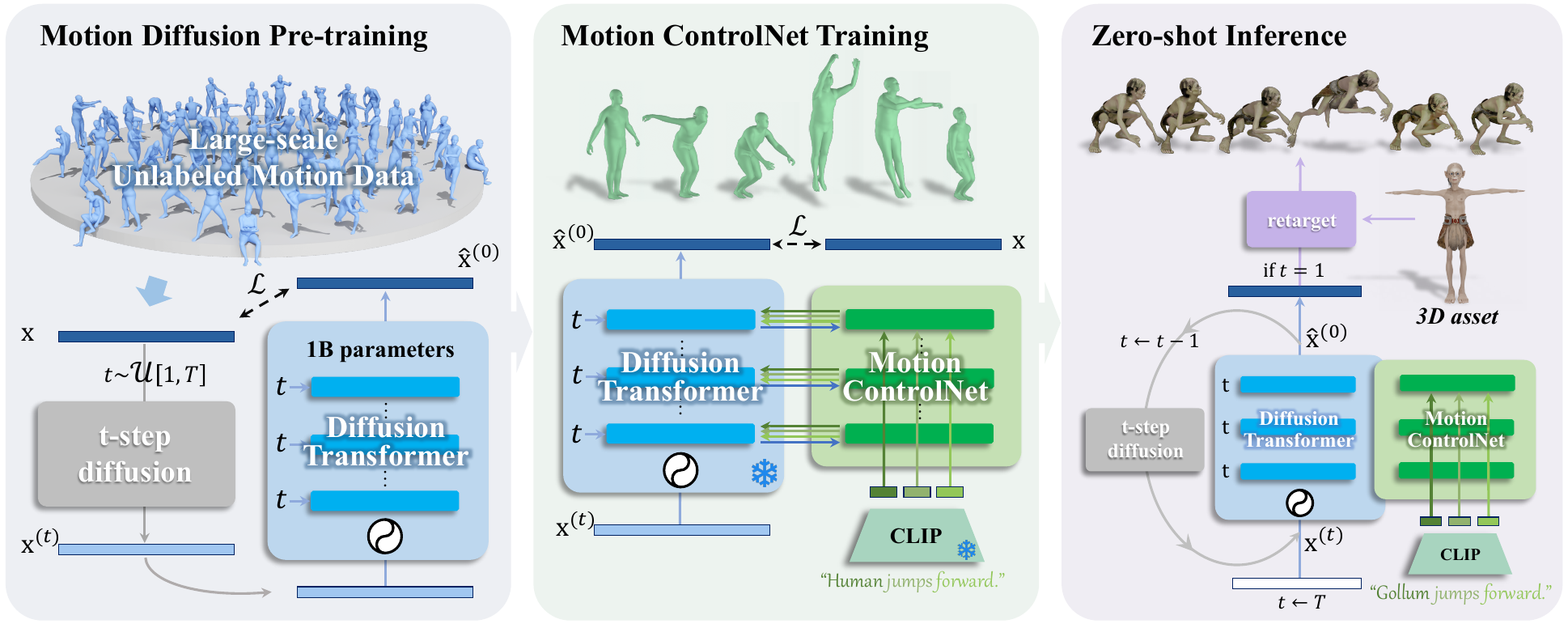} 
	\caption{\textbf{Method overview}. 
 We train our OMG model in two stages. First, we leverage large-scale unlabeled motion data to pre-train an unconditional diffusion model with up to 1B parameters (\cref{sec:3.1}). Then, we adopt a conditional fine-tuning scheme called motion ControlNet to condition the pre-trained diffusion model on text prompts (\cref{sec:3.2}).
 During inference, the pre-trained unconditional denoiser and the fine-tuned conditional denoiser are combined with classifier-free guidance, generating realistic motions with zero-shot text inputs.
 } 
	\label{fig:overview} 
\end{figure*}

\myparagraph{Conditional Motion Synthesis.}
Being able to generate realistic and contextually relevant motion sequences based on various types of conditions, conditional motion synthesis has received increasing attention in the field of motion generation. Common types of conditions include text~\cite{lin2018human, ahn2018text2action, ahuja2019language2pose, tevet2022motionclip, petrovich2022temos, yuan2023physdiff, chen2023executing, guo2022generating, kim2023flame, shafir2023human, liang2023intergen}, action~\cite{guo2020action2motion, petrovich2021action}, music~\cite{lee2019dancing, aggarwal2021dance2music, li2021ai, li2022danceformer}, speech~\cite{habibie2022motion, ao2022rhythmic, ao2023gesturediffuclip, zhao2024media2face}, control signals~\cite{starke2019neural, starke2022deepphase, peng2021amp, xie2023omnicontrol}, action labels~\cite{guo2020action2motion, petrovich2021action, xu2023actformer}, incomplete motion sequences~\cite{duan2021single, harvey2020robust}, images~\cite{rempe2021humor, chen2022learning, he2021challencap, liang2023hybridcap, ren2023lidar} and scene~\cite{Zhang_2020_CVPR}.
The advent of Diffusion Models~\cite{ho2020denoising, song2020score} has given a big boost to text-driven motion synthesis.~\citet{kim2023flame} develops a transformer-based diffusion model which could generate and edit motions well aligned with the given text. Motion Diffusion Model (MDM)~\cite{tevet2022human} combines insights already well established in the motion generation domain and in each diffusion step predicts the sample rather than the noise. Motion Latent-based Diffusion model (MLD)~\cite{chen2023executing} performs a diffusion process on the motion latent space. Besides the diffusion model, T2M-GPT~\cite{zhang2023t2m} investigates a framework that combines VQ-VAE~\cite{van2017neural} and autoregressive model. However, since these models are trained only based on paired text-motion datasets such as HumanML3D~\cite{guo2022generating}, they cannot well handle unseen text prompts. 

\myparagraph{Open-vocabulary Generation.}
Instead of relying on pre-existing data, zero-shot text-driven generation leverages general knowledge learned during training to create novel content from text prompts. \citet{reed2016generative} uses DC-GAN architecture to steer for zero-shot text-to-image synthesis. CLIP~\cite{radford2021learning}, pre-trained on four hundred million image-text pairs, possesses the remarkable ability to effectively comprehend and generate meaningful associations between images and text. With CLIP's strong ability, many works are able to generate high-quality zero-shot text-driven images~\cite{frans2022clipdraw, xu2022simple, patashnik2021styleclip} or 3D objects~\cite{jain2022zero, michel2022text2mesh, sanghi2022clip, peng2021animatable, jetchev2021clipmatrix, wang2022clip}.
In the motion synthesis field, some works have attempted to investigate open-vocabulary text-to-motion generation and achieved good results. MotionCLIP~\cite{tevet2022motionclip} utilizes a decoder to decode the CLIP motion embeddings. AvatarCLIP~\cite{hong2022avatarclip} synthesizes a key pose and then retrieves the closest motion from the database. \citet{lin2023being} pre-train a motion generator that learns to reconstruct the full motion from the masked motion with the key pose. Make-An-Animation~\cite{azadi2023make} pre-trains generative model with diverse in-the-wild text-pose pair. However, since these approaches are based on text-pose alignment, the generated motion is deficient in realism due to a lack of global rotation and translation. Recently, MotionGPT~\cite{jiang2023motiongpt} pre-trains and fine-tunes the large language model with tokenized motion and text data. However, it still struggles to generate novel motion from open-vocabulary text prompts.

\myparagraph{Mixture-of-Experts.} In the deep learning field, Mixture-of-Experts (MoE)~\cite{jacobs1991adaptive, shazeer2017outrageously, fedus2022switch} in a neural network architecture divides specific model parameters into distinct groups, referred to as ``experts''. 
\citet{eigen2013learning} use a different gating network at each layer in a multi-layer network, increasing the number of effective experts by introducing an exponential number of paths through different combinations of experts at each layer.
Other research that regards the entire model as an expert~\cite{li2022branch} also achieves good results. Moreover, endeavors are underway to enhance the training and inference speed within the MoE framework~\cite{he2021fastmoe, lepikhin2020gshard}.
The MoE paradigm is applicable not only to language processing tasks but also extends to visual models~\cite{riquelme2021scaling}, Mixture-of-Modality-Experts in multi-modal transformers~\cite{shen2023scaling}, as well as motion synthesis.~\citet{holden2017phase} develop phase-functioned experts blended by pre-defined blending weights to control character at a specific phase, and~\citet{starke2022deepphase, starke2020local, zhang2018mode} uses a gating network to predict the blending weights, achieving impressive results of character control.  Inspired by them, we devise text-token-specific experts to control the corresponding sub-motions.
\section{Methods}

We aim to generate realistic and diverse human motions that are conditioned on text prompts, which capture complex and abstract motion characteristics with zero-shot open-vocabulary descriptions. %
To this end, we adopt pretrain-then-finetune paradigm to enhance the capability of our model. 
For the unconditional denoiser (\cref{fig:overview} left), we leverage a large-scale unlabeled motion dataset for motion diffusion pre-training and scale the model size up (\cref{sec:3.1}). 
For the conditional denoiser  (\cref{fig:overview} middle), we devise a specific conditional fine-tuning scheme called motion ControlNet, including a novel conditioning design called Mixture-of-Controllers, to exploit the zero-shot capability of the CLIP text embeddings (\cref{sec:3.2}). 
During inference (\cref{fig:overview} right), we further use classifier-free guidance to combine the unconditional denoiser and the conditional one.

\subsection{Motion Diffusion Pre-training} 
\label{sec:3.1}

\myparagraph{Large-scale Unlabeled Motion Dataset.}
To facilitate large-scale pre-training in the motion domain, we collected a large amount of high-quality human motion data totaling over 20M frames from various sources, such as character animation datasets, marker-based, IMU-based, and multi-view markerless mocap datasets.
We then standardize the frame rate and re-target them to the unified parametric skeleton of SMPL body model~\cite{loper2023smpl} using off-the-shelf re-targeting algorithms~\cite{zhang2023skinned}. 
After that, following the previous successful works~\cite{Guo_2022_CVPR, tevet2022human} on motion generation, we enrich the SMPL skeletons with an over-parameterized motion representation that facilitates network learning.

\begin{table}[tbp]

    \centering
    \resizebox{0.98\linewidth}{!}{
        \begin{tabular}{@{}lccccc@{}}
            \toprule
            Model Name & $n_{layers}$ & $d_{model}$ & $n_{heads}$ & $d_{head}$   & $n_{params}$  \\
            \midrule
              OMG-Base &
               8 &
               1024 &
               8 &
               128&
               88M
               \\
              OMG-Large &
               12 &
               1280 &
               10 &
               128&
               201M
               \\
               OMG-Huge &
               16 &
               1664 &
               13 &
               128&
               405M
               \\
               OMG-Giant &
               24 &
               2048 &
               16 &
               128&
               1B
               \\
            \bottomrule
        \end{tabular}
        
    }
    \caption{Sizes and architectures of our 4 models.}
    \label{tab:1}
\end{table}

\myparagraph{Model Scaling Up.}
We design our pre-trained diffusion model with minimalism, only preserving the essential modules that are scalable and efficient to adapt to expanding data.
To this end, we adopt the Diffusion Transformer (DiT)~\cite{peebles2023scalable} architecture as our network backbone, due to its scalability and impressive performance with increasing input tokens and training data.
The only difference is that we use rotary positional embedding~\cite{su2021roformer} to encourage the model to capture the relative temporal relations among the frames.
To study the dependence of performance on model size, we train 4 different sizes of our OMG model, ranging from 88 million parameters to 1 billion parameters.
\cref{tab:1} shows the configuration details of our 4 models. 
Here $n_{layers}$ is the total number of layers, $d_{model}$ is the number of units in each bottleneck layer (we always have the feed-forward layer two times the size of the bottleneck layer, $d_{ff}$ = 2$d_{model}$), and $d_{head}$ is the dimension of each attention head, and $n_{params}$ is the total number of trainable parameters.

\myparagraph{Training.}
We train our unconditional denoiser $\mathcal{D}_{u}$ to predict the clean motion $\mathbf{\hat{x}^{(0)}}$ given diffusion timestep $t$ with the simple objective:
\begin{align}
\mathcal{L}_{simple} = 
\mathbb{E}_{\mathbf{x}, t, \epsilon} [&\lambda_t  ||\mathbf{x} -\mathcal{D}_{u}(\mathbf{x}^{(t)}, t)||_2^2 ],
\end{align}
where $\mathbf{x}^{(t)}=\mathbf{x}+\sigma_t \epsilon$ is $t$-step noised motion, noise $\epsilon \sim \mathcal{N}(\mathbf{0},\mathbf{I})$, $\lambda_t$ is the loss weighting factor.
Furthermore, we introduce geometric losses, analogous to MDM~\cite{tevet2022human}, that contain velocity loss $\mathcal{L}_{vel}$, and foot contact loss $\mathcal{L}_{foot}$ to enforce physical properties and anti-sliding. Overall, the unconditional denoiser $\mathcal{D}_{u}$ is thus trained with the following objective:
\begin{align}
\label{eq:loss}
\mathcal{L} = \mathcal{L}_{simple} + \mathcal{\lambda}_{vel} \mathcal{L}_{vel} + \mathcal{\lambda}_{foot} \mathcal{L}_{foot}.
\end{align}

In our experiments, $\mathcal{\lambda}_{vel}=30$ and $\mathcal{\lambda}_{foot}=30$. The number of diffusion timesteps $T$ is set to 1,000 with cosine noise level schedule~\cite {nichol2021improved}.
We train all sizes of the model for 1M iterations with a batch size of $256$ using the AdamW~\cite{loshchilov2017decoupled} optimizer with a weight decay of $0$, a maximum learning rate of $10^{-4}$, and a cosine LR schedule with 10K linear warmup steps. 
The models are trained using Pytorch with ZeRO~\cite{rajbhandari2020zero} memory redundancy strategy on 8 NVIDIA A100 GPUs. The largest model OMG-G takes 1500 GPU hours for pre-training.

\begin{figure}[tbp] 
	\centering  
 
	\includegraphics[width=1\linewidth]{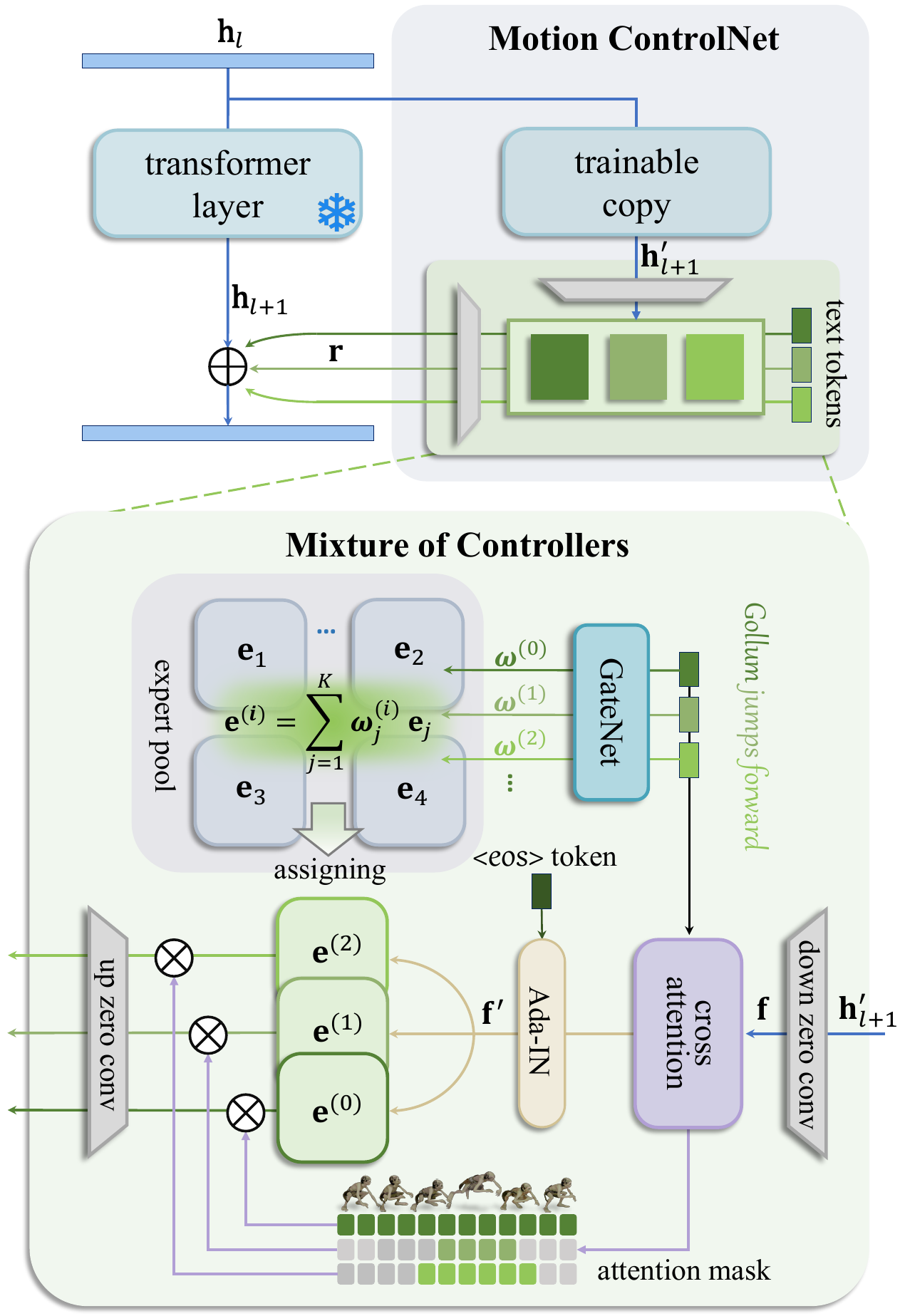} 
	\caption{\textbf{Motion ControlNet (top)} freezes the parameters of the pre-trained transformer layer and combines a trainable copy of the layer with the proposed \textbf{Mixture-of-Controllers (bottom)} block. The MoC block first fuses the text features and motion features and simultaneously determines the sub-motion ranges for each text token with the cross-attention mechanism. Then it performs fine-grained control of sub-motions with text-token-specific experts. 
 } 
	\label{fig:control} 
 \vspace{-8pt}
\end{figure}

\begin{figure*}[htbp] 
	\centering  
	\includegraphics[width=\linewidth]{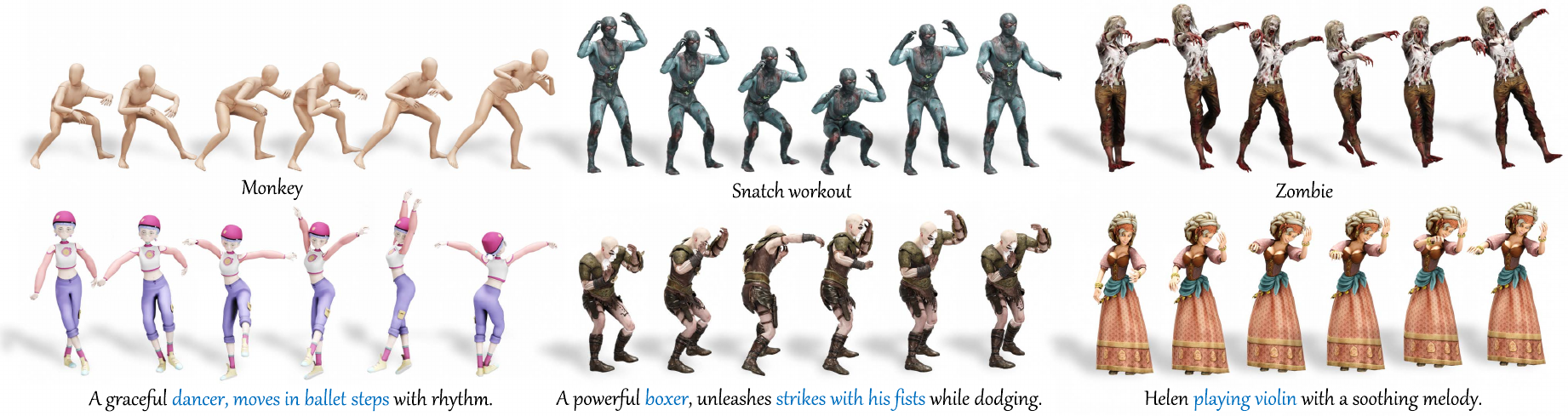} 
	\caption{\textbf{Qualitative results} generated by our model given various unseen text prompts. Our model effectively captures the motion characteristics from either a single phrase or longer natural sentences.} 
	\label{gallery} 
\end{figure*}

\myparagraph{Sliding Random Windows.}
During training, we propose sliding random windows that iterate over each motion state frame $x_i$ in the motion dataset to sample motion sequences starting from $x_i$. 
Specifically, for the $i^{th}$ frame, we randomly crop a sub-sequence $\mathbf{x}$ of length $l$ with a random window $[i, i+l)$, where $l\sim \mathcal{U}[1, L]$ is a uniform variable, and $L$ is the hyper-parameter that determines the maximum length of the generated motion sequence of our model. 
This pushes the model to learn the relations among both temporal frames and spatial features of a single motion state to process arbitrary lengths of motions even a single keyframe, to facilitate the following fine-grained control of sub-motions of arbitrary lengths. 
In our experiments, for balancing resolution and duration, the maximum length $L$ of sliding random windows is set to $300$ with the framerate set to $30$.

\subsection{Motion ControlNet} 
\label{sec:3.2}

To incorporate the text prompts $\mathbf{c}=\{\mathbf{c}_i\}_{i=1}^n$ into the pre-trained motion model,
We adopt a fine-tuning scheme we call motion ControlNet to condition on text prompts, inspired by image ControlNet~\cite{zhang2023adding}, including a trainable copy of original transformer layers and the proposed conditioning blocks called Mixture-of-Controllers (MoC).

As illustrated in \cref{fig:control} top, we freeze the parameters of the pre-trained transformer layer and combine a trainable copy of the layer with a MoC block that injects $n$ residuals corresponding to $n$ text tokens into the output of the original transformer layer. 
Specifically, the frozen parameters retain the pre-trained denoising ability, while the trainable copy reuses the large-scale pre-trained model as a strong initial backbone to learn to extract semantic motion features from intermediate representation $\mathbf{h}_l$ of the last layer. 
And based on that, the MoC block is employed to predict the $n$ conditional residuals $\mathbf{r}$ corresponding to $n$ text tokens, all of which are then added to the original output $\mathbf{h}_{l+1}$.
To do so, a pre-trained CLIP text encoder $\mathcal{E}(\mathbf{c})$ is employed to extract the text token embeddings.

\begin{figure*}[htb] 
	\centering  
	\includegraphics[width=\linewidth]{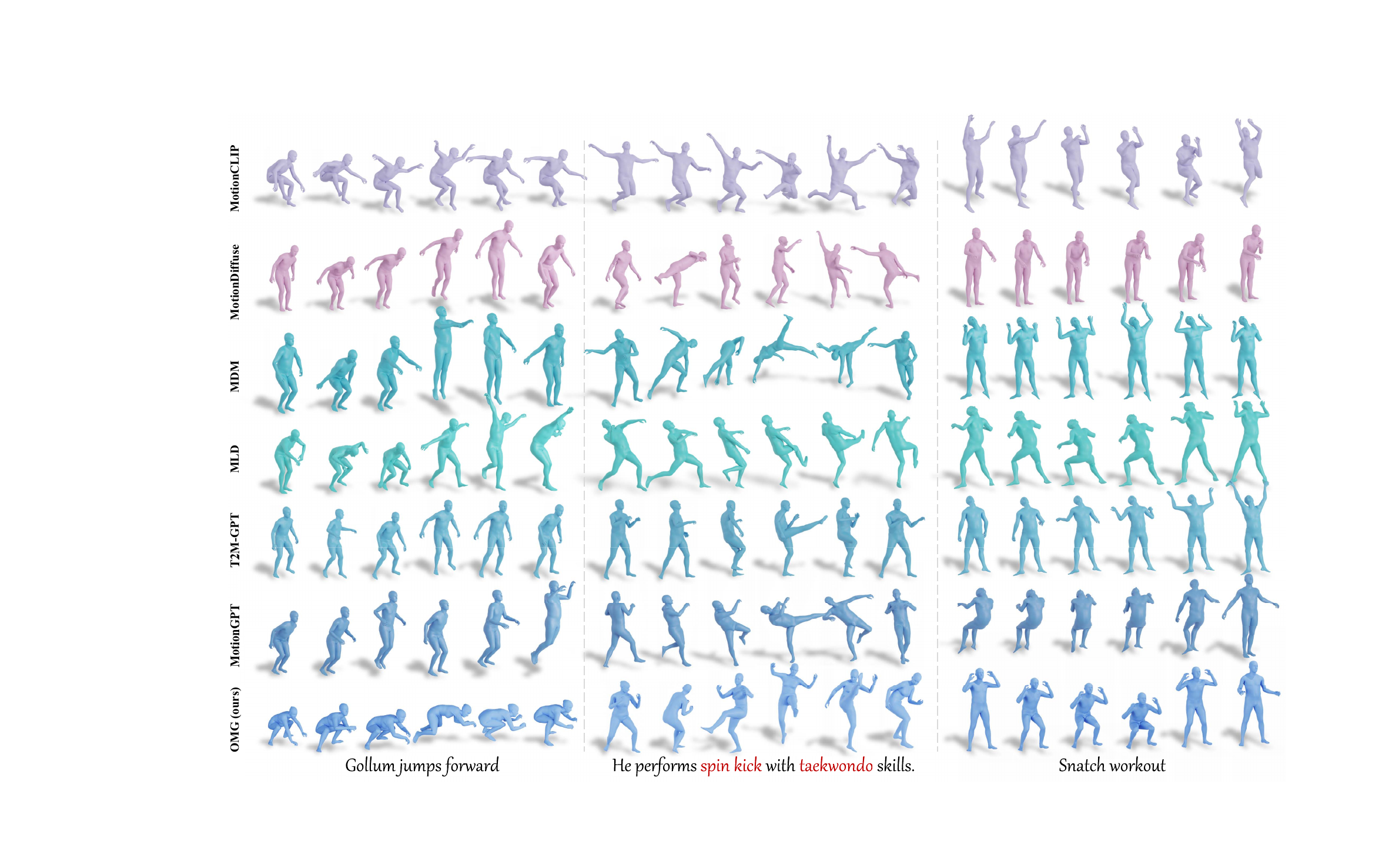} 
	\caption{\textbf{Qualitative comparison.} Our method can generate high-quality human motions that better align with text prompts than previous state-of-the-art methods. } 
	\label{comp} 
 \vspace{-4pt}
\end{figure*}

\myparagraph{Mixture-of-Controllers.}
In the Mixture-of-Controllers block (\cref{fig:control} bottom), our key idea is to control the sub-motion sequences separately with different expert controllers in an MoE fashion~\cite{shazeer2017outrageously,fedus2022switch}, so as to better align them to the corresponding text token embeddings in CLIP space. 
For instance, for the given text prompt \emph{``Gollum jumps forward''}, the \emph{``Gollum''} token corresponds to the entire sequence while \emph{``jumps''} corresponds only to the sub-sequence when he is jumping. 

To this end, we introduce two cooperative designs.
The first one is a \textbf{cross-attention mechanism} to fuse the text features and motion features and simultaneously determine the sub-motion ranges for each text token.
The second one is the \textbf{text-token-specific experts} selected from an expert pool to perform fine-grained control of sub-motions.

For the \textbf{cross-attention mechanism}, a sequence of motion features $\mathbf{f}\in \mathbb{R}^{l\times d_m}$ is input into a cross-attention layer along with text embeddings $\mathcal{E}(\mathbf{c}) \in \mathbb{R}^{n\times d_c}$, where $l$ denotes the length of the sequence, $d_m$ represents the dimension of the motion features, $n$ denotes the number of the text tokens, and $d_c$ represents the dimension of text embedding.
The cross-attention layer projects $\mathbf{f}$ to a query matrix $\textbf Q = \mathbf{f} \cdot \mathbf{W}_m^{Q}$, where $W_m^{Q} \in \mathbb{R}^{d_m\times d_m}$ is the motion projection matrix.
And $\mathcal{E}(\mathbf{c})$ is projected to a key matrix $\textbf K = \mathcal{E}(\mathbf{c}) \mathbf{W}_c^{K}$ and a value matrix $\textbf V = \mathcal{E}(\mathbf{c}) \mathbf{W}_c^{V}$, where $W_c^{K}, W_c^{V} \in \mathbb{R}^{d_c\times d_m}$ are the text projection matrices.
Then the text values $\textbf V$ are distributed into the motion sequence with the attention mechanism:
\begin{align}
\mathbf{f}' = \mathbf{f} + \text{softmax}(\frac{\mathbf{Q}\mathbf{K}^T}{\sqrt{d}})\mathbf{V}.
\end{align}
Simultaneously an attention map between text tokens and motion sequence is produced, denoted by $\mathbf{A}= \text{softmax}(\frac{\mathbf{Q}\mathbf{K}^T}{\sqrt{d}}) \in \mathbb{R}^{l\times n}$.
Within this attention map, the $i^{th}$ column $\mathbf{A}_{\ast,i}$ indicates a correspondence between a sub-motion sequence and the $i^{th}$ text token.
Besides, we employ the adaptive instance normalization (Ada-IN) conditioning on $\langle eos\rangle$ (end of sequence) token to normalize $\mathbf{f}'$ before sending them to the experts.
Since CLIP $\langle eos\rangle$ token summarizes the entire text, we intuitively use it to unify the distribution of the entire motion sequence.

For the \textbf{text-token-specific experts}, we define an expert controller corresponding to $i^{th}$ text token $\mathcal{E}(\mathbf{c}_i)$ as a two-layer feed-forward network (denoted by $\mathcal{F}$) with parameters $\mathbf{e}^{(i)}=\{\mathbf{W}_0\in \mathbb{R}^{d_m\times 2d_m},  \mathbf{W}_1\in \mathbb{R}^{2d_m\times d_m}, \mathbf{b}_0\in \mathbb{R}^{2d_m}, \mathbf{b}_1\in \mathbb{R}^{d_m}\}$.
Furthermore, the parameters $\mathbf{e}^{(i)}$ are generated by blending $K$ expert parameters $\{\mathbf{e}_1,...,\mathbf{e}_K\}$ in an expert pool, each of which is in a form of the same parameter configuration:
\begin{align}
\mathbf{e}^{(i)} = \sum_j^K \bm{\omega}_j^{(i)} \mathbf{e}_j,
\end{align}
where $j$ indicates the index of expert pool, $K$ is expert pool size that can be adjusted, and $\bm{\omega}^{(i)}=\{\bm{\omega}^{(i)}_j\}_{j=1}^{K}\in \mathbb{R}^K$ is the blending weights vector controlled by the text token $\mathbf{c}_i$ with a three-layer fully-connected gating network $\mathcal{G}$:
\begin{align}
\bm{\omega}^{(i)} = \text{softmax}(\mathcal{G}(\mathcal{E}(\mathbf{c}_i))).
\end{align}
Then, the $n$ experts take motion features $\mathbf{f}'$ as input and output $n$ conditional masked residuals $\mathbf{r}=\{\mathbf{r}_i\}_{i=1}^{n}$:

\begin{align}
\mathbf{r}_i = \mathbf{M}_{\ast, i}\circ \mathcal{F}(\mathbf{f}'|\mathbf{e}^{(i)}),  
\end{align}
where $\mathbf{M}_{\ast, i}=\text{sigmoid}(\gamma(\mathbf{A}_{\ast, i}-\beta \text{max}(\mathbf{A}_{\ast, i})))$ is an attention mask ranging from 0 to 1, and $\gamma$ and $\beta$ are hyper-parameters to control the sharpness and threshold respectively.

Additionally, a down-projection and up-projection 1-D convolution pair is used to process input and output respectively to reduce the control latent dimension and thus the trainable parameters.
Moreover, similar to image ControlNet~\cite{zhang2023adding}, we use zero-initialized convolution parameters to protect the trainable copy from the harmful gradient noises in the initial training steps.

\myparagraph{Training and Inference.}
The conditional denoiser $\mathcal{D}_{c}$ is also trained using the same objective $\mathcal{L}$ (\cref{eq:loss}) with trainable parameters in motion ControlNet.
In our experiments, we employ the frozen \emph{CLIP-VIT-L/14}~\cite{radford2021learning} text encoder to extract text embeddings at the final layer normalization with dimension $d_c=768$ and we truncate the text tokens with the maximum token number $77$.
We set the down-projection latent dimension $d_m$ of MoC blocks to $256$, the expert pool size $K=12$, and attention mask sharpness $\gamma=24$ and threshold $\beta=0.25$.
We train all the motion ControlNets for $500$ epochs with a batch size of $64$ using the AdamW optimizer with a weight decay of $1\times 10^{-5}$, a maximum learning rate of $3\times 10^{-5}$, and a cosine LR schedule with 1K linear warmup steps.
In the training process, we randomly replace $\langle eos\rangle$ token with the empty token with 50\% probability to increase the ability to capture global semantics in text tokens as a replacement.

During inference, classifier-free guidance is used to combine unconditional denoiser and conditional denoiser: 
\begin{align}
\hat{\mathbf{x}}^{(0)} = (1-s)\cdot \mathcal{D}_{u}(\mathbf{x}^{(t)}, t) + s \cdot \mathcal{D}_{c}(\mathbf{x}^{(t)}, t, \mathbf{c}).
\end{align}
In our experiments, DDIM~\cite{song2020denoising} sampling strategy with 200 timesteps is employed and the guidance strength $s=4.5$.

\section{Experiments}
\label{sec:exp}


We show qualitative results in \cref{gallery}. Our method enables fine-grained control of complex and abstract motion trait descriptions.
We encourage the reader to appreciate more qualitative results provided in supplements and the video.

\begin{table}[tb]

	\begin{center}
		\centering

		\resizebox{1\linewidth}{!}{
    		\begin{tabular}{@{}lccccccc@{}}
    		    \toprule
    		    \multirow{2}{*}{Methods} & \multicolumn{3}{c}{HumanML3D~\cite{guo2022generating}} & \multicolumn{3}{c}{Mixamo~\cite{Mixamo} (zero-shot)}
                \\  \cmidrule(lr){2-4} \cmidrule(lr){5-7}
    			& FID$\downarrow$ &R-Precision$\uparrow$ & Diversity $\rightarrow$& FID $\downarrow$   & CLIP-score$\uparrow$&  Diversity $\rightarrow$ \\
        
    			\toprule
                   Real & 
                  $0.002^{\pm.000}$ &
                  $0.797^{\pm.002}$ &
                  $9.503^{\pm.065}$ &
                  $0.106^{\pm.003}$ &
                  $0.648^{\pm.001}$ &
                  $2.665^{\pm.022}$
                  \\  \midrule
                
                MotionCLIP~\cite{tevet2022motionclip}& 
                -&-&-&
                $2.542^{\pm.012}$ &
                $0.511^{\pm.004}$ &
                $2.205^{\pm.012}$
                \\
                MAA~\cite{azadi2023make}& 
                $0.774^{\pm.002}$ &
                $0.676^{\pm.001}$ &
                $8.230^{\pm.064}$ &
                -&-&-
                
                \\
                MotionDiffuse~\cite{zhang2022motiondiffuse}& 
                $0.630^{\pm.001}$ &
                $\underline{0.782}^{\pm.001}$ &
                $9.410^{\pm.049}$ &
                $2.363^{\pm.010}$ &
                $0.505^{\pm.002}$ &
                $2.411^{\pm.016}$
                \\
                
    			MDM~\cite{tevet2022human}&
                $0.544^{\pm.044}$ &
                $0.611^{\pm.007}$ &
                $\underline{9.559}^{\pm.086}$ &
                $1.297^{\pm.004}$ &
                $0.536^{\pm.004}$ &
                $\underline{2.594}^{\pm.011}$
                \\
                MLD~\cite{chen2023executing} &
                $0.473^{\pm.013}$ &
                $0.772^{\pm.002}$ &
                $9.724^{\pm.082}$ &
                $\underline{1.229}^{\pm.004}$ &
                $\underline{0.556}^{\pm.003}$ &
                $2.583^{\pm.018}$
                \\
                
                T2M-GPT~\cite{zhang2023t2m} &
                $\textbf{0.116}^{\pm.004}$ &
                $0.775^{\pm.002}$ &
                $9.844^{\pm.095}$ &
                $1.420^{\pm.003}$ &
                $0.541^{\pm.002}$ & 
                $2.590^{\pm.022}$
                \\
                MotionGPT~\cite{jiang2023motiongpt}&
                $\underline{0.232}^{\pm.008}$ &
                $0.778^{\pm.002}$ &
                $\textbf{9.528}^{\pm.071}$ &
                $1.365^{\pm.003}$ &
                $0.552^{\pm.002}$ & 
                $2.589^{\pm.018}$
                \\
                
                \midrule
                OMG (ours) & 
                $0.381^{\pm.008}$ &
                $\textbf{0.784}^{\pm.002}$ &
                $9.657^{\pm.085}$ &
                $\textbf{1.164}^{\pm.009}$ &
                $\textbf{0.588}^{\pm.002}$ & 
                $\textbf{2.632}^{\pm.021}$
                \\
    			\bottomrule
    		\end{tabular}
		}
    \caption{Comparison of text-to-motion generation on HumanML3D~\cite{guo2022generating} and Mixamo~\cite{Mixamo} test set. We ran all the evaluations 20 times, with the average reported alongside a 95\% confidence interval. The right arrow $\rightarrow$ means
    the closer to real motion the better. \textbf{Bold} and \underline{underline} indicate
    the best and the second best result. The term (Zero-shot) implies that the dataset contains unseen open-vocabulary texts.
    }
    \label{tab:Comparison}
	\end{center}
\end{table}

\subsection{Datasets and Evaluation Metrics}
\label{sec:4.2}
\myparagraph{Training Datasets.} At the pre-training stage, we utilize various publicly available human motion datasets, such as artist-created datasets~\cite{harvey2020robust,mason2022local}, marker-based ~\cite{mahmood2019amass,taheri2020grab,araujo2023circle,liu2022beat,ionescu2013human3}, IMU-based~\cite{liang2023hybridcap,trumble2017total} and multi-view markerless~\cite{zhang2022egobody, li2021ai, cai2022humman, liang2023intergen} motion capture datasets, totaling over 20 million frames.  In the subsequent conditional fine-tuning stage, we train our motion ControlNet using the text-motion HumanML3D~\cite{guo2022generating} dataset, for fair comparisons with previous methods.

\myparagraph{Evaluation Datasets.} We test on two benchmarks, the HumanML3D~\cite{guo2022generating}  and the Mixamo~\cite{Mixamo} test set. The HumanML3D test set evaluates the in-domain performance. And the Mixamo dataset consists of abundant artist-created animations and human-annotated descriptions, offering a wide variety of diverse and dynamic motions, utilized to compare the zero-shot performance across the domains.

\myparagraph{Evaluation Metrics} are summarized as four parts.
(1) Frechet Inception Distance (FID) is our principal metric to evaluate the feature distributions between the generated and real motions. 
(2) Motion-retrieval precision (R-Precision) calculates the text and motion Top 3 matching accuracy under feature space.
(3) CLIP-score. Borrowing from text-to-image synthesis~\cite{saharia2022photorealistic}, we use CLIP-score to evaluate zero-shot text-motion consistency by measuring cosine similarity in CLIP space. To do so, we train a motion encoder using the training set of HumanML3D and Mixamo to extract motion features that are aligned to text embeddings in CLIP space as same as~\cite{tevet2022motionclip}.
(4) Diversity is assessed by calculating variance through features.

\subsection{Comparison}
\label{sec:4.3}
We compare our approach with various state-of-the-art methods. Specifically, we apply conditional motion diffusion models, including MDM~\cite{tevet2022human}, MLD~\cite{chen2023executing}, MotionDiffuse~\cite{zhang2022motiondiffuse}, VAE-based model MotionCLIP~\cite{tevet2022motionclip}, and auto-regressive model T2M-GPT~\cite{zhang2023t2m}. Besides, we also apply MAA~\cite{azadi2023make} based on the text-pose alignment method, and MotionGPT~\cite{jiang2023motiongpt} utilizing motion-language pre-training. Some results on HumanML3D are borrowed from their own paper.
The quantitative results are presented in \cref{tab:Comparison}. Our method demonstrates the best text-to-motion alignment (R-Precision) in the in-domain evaluation. Moreover, among all diffusion-based methods, our model achieves the best FID. In terms of zero-shot performance, our approach achieves the best FID and CLIP-score, outperforming previous methods. This suggests a superior capability for high-quality motion generation and effective matching with zero-shot text prompts. As depicted in \cref{comp} of the qualitative results, our model demonstrates the capability to generate motions that show more realism, and better alignment with each motion characteristic description, whether in the sentence or phrase.

\begin{figure}[tb] 
	\centering  
	\includegraphics[width=\linewidth]{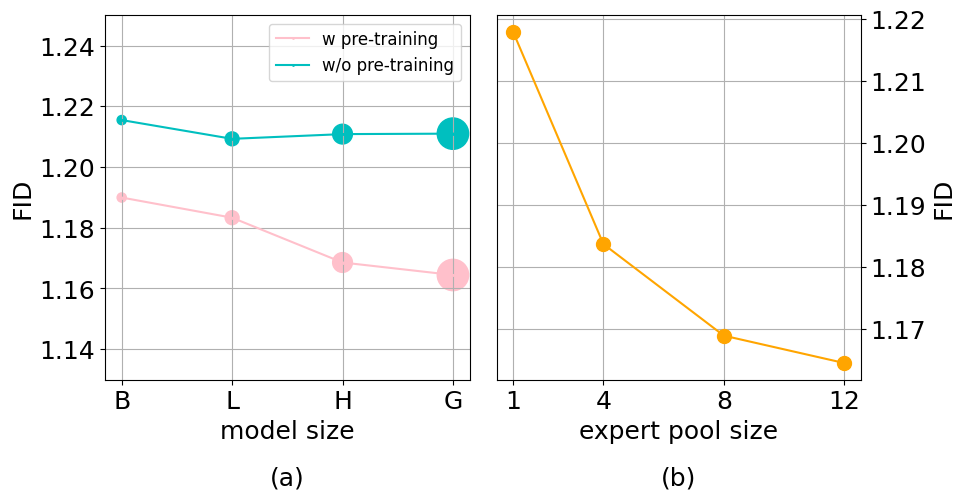} 
	\caption{Quantitative evaluation on pre-training, model size, and expert pool size. (a) Models \emph{w\_pre-training} show consistently improved performance over \emph{w/o\_pre-training}, and \emph{w\_pre-training} models, which benefit from large-scale motion data, improve with increasing model size. (b) Larger expert pool sizes improve the performance.} 
	\label{fig:curve} 
 \vspace{4pt}
\end{figure}

\subsection{Ablation Study}
\label{sec:4.4}
To examine the specific contributions of our novel OMG model architecture, we conduct a series of ablation studies focusing on the roles of pre-training, model scale, and multiple expert controllers. Additionally, we undertake an in-depth analysis of our MoC Block. These evaluations utilize the out-of-domain dataset Mixamo~\cite{Mixamo}, providing a robust testbed to ascertain the effectiveness of the technical designs on the zero-shot performance of our model.

\begin{figure}[tb] 
	\centering  
	\includegraphics[width=1.02\linewidth]{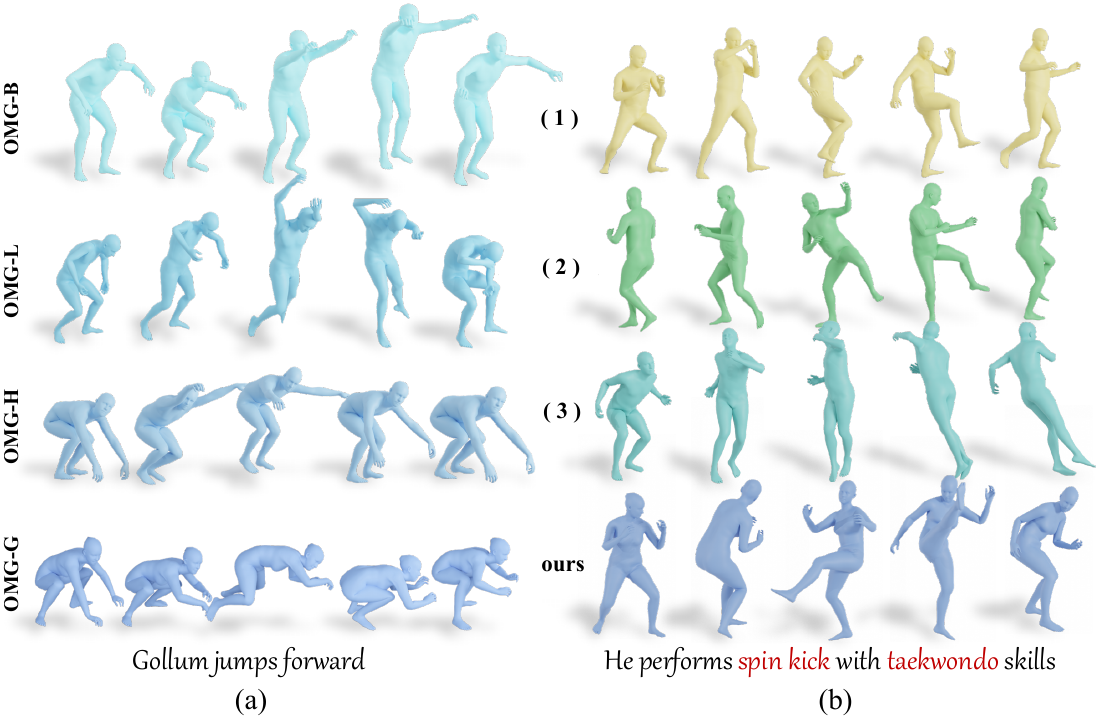} 
	\caption{Qualitative evaluation on model sizes (a) and MoC block (b). Models with larger sizes effectively comprehend richer out-of-domain motion features to present better motion expressions. Besides, our technical designs effectively improve the alignment with the input texts.} 
	\label{gallery1} 
 \vspace{4pt}
\end{figure}

\myparagraph{Effect of Pre-training and Model Scale.}
Our investigation into the impact of pre-training involves training several variant models without pre-training (\textit{w/o\_pre-training}) across four distinct scales (see \cref{tab:1}). These models are then compared against counterparts with the pre-training process (\textit{w\_pre-training}). As illustrated in \cref{fig:curve}a, \textit{w\_pre-training} consistently outperforms \textit{w/o\_pre-training}, achieving lower FID across all sizes. This indicates that our model's performance on zero-shot motion generation is effectively enhanced through pre-training. Meanwhile, we observe that the performance of \textit{w\_pre-training} improves with increasing model size. Qualitative analysis, as shown in \cref{gallery1}a, reveals that the \emph{``Gollum''} characteristic of the generated motions becomes more pronounced with larger models. This suggests that larger model sizes 
enhance the overall quality and alignment of the generated motions.

\myparagraph{Effect of Expert Pool Size.}
To systematically analyze the impact of multiple experts, we train four distinct variants of the OMG model, sweeping over different expert pool sizes ($K=1, 4, 8$, and $12$ respectively). The results, as depicted in \cref{fig:curve}b, demonstrate a discernible decrease in FID as expert pool size increases. This trend suggests that the larger expert hypothesis space enhances the experts' ability to align sub-motions with their respective text embeddings.

\begin{table}[tb]
    \centering
    \resizebox{1\linewidth}{!}{
        \begin{tabular}{@{}lcccc@{}}
            \toprule
            Methods & FID $\downarrow$ & CLIP-score $\uparrow$ & Diversity $\rightarrow$ \\  
            \midrule
            Real & 
            $0.106^{\pm.003}$ &
            $0.648^{\pm.001}$ &
            $2.665^{\pm.022}$\\  
            \midrule
            (1) Cross-attn + FFN &
            $1.252^{\pm.006}$ &
            $0.552^{\pm.003}$ & 
            $2.576^{\pm.025}$\\
            (2) w/o Zero Conv &
            $1.339^{\pm.005}$ &
            $0.535^{\pm.004}$ & 
            $2.695^{\pm.014}$\\
            (3) w/o Attention Mask&
            $1.246^{\pm.008}$ &
            $0.557^{\pm.003}$ & 
            $\textbf{2.647}^{\pm.026}$\\
            \midrule
            ours (complete) &
            $\textbf{1.164}^{\pm.009}$ &
            $\textbf{0.588}^{\pm.002}$ & 
            $2.632^{\pm.021}$\\
            \bottomrule
        \end{tabular}
    }
    \caption{
    Quantitative evaluation on MoC block. The damping performance of the three variants of our model highlights the effectiveness of our MoC block technical designs.
    }
    \vspace{4pt}
    \label{tab:MoCblock}
\end{table}

\myparagraph{Evaluation of the MoC Block.}
In order to assess the effectiveness of our proposed MoC block, we explore its performance through several variant configurations: \textbf{(1)} Cross-attn + FFN, where the architecture is pruned to include only a cross-attention layer and a feed-forward network; \textbf{(2)} w/o Zero Conv, substituting zero convolutions with standard convolutional layers initialized using Gaussian distributions; and \textbf{(3)} w/o Attention Mask, omitting the multiplication of attention masks with the output of the expert layer. 
These variant models are rigorously compared against our original OMG model.
Quantitative results, as shown in \cref{tab:MoCblock}, demonstrate that the exclusion of specific components within the MoC Block leads to a worse result in both the FID and CLIP-score. This observation underscores the integral contributions of these components to the motion generation process. Qualitative analysis, presented in \cref{gallery1}b, indicates noticeable deficiencies in the motion generation of all three variants: variant \textbf{(1)} is not able to clearly depict \emph{``spin''} motions; variant \textbf{(2)} exhibits less expressive motion features; and variant \textbf{(3)} inadequately captures the \emph{``kick''} motion. In contrast, our OMG model demonstrates a superior ability to more accurately align motion with the text prompt.

\section{Conclusion}
\label{sec:discussion}

In this paper, we present a novel text-to-motion generation framework, OMG, that combines the advantages of conditional generative models and text-pose alignment methods. It carefully tailors pretrain-then-finetune paradigm into text-to-motion generation. 
The pre-training stage leverages a large amount of unlabeled motion data to train a powerful unconditional diffusion model that ensures the realism and diversity of the generated motions. 
The fine-tuning stage introduces motion ControlNet including the proposed novel text conditioning block called Mixture-of-Controllers. With the cross-attention mechanism and text-token-specific experts, it adaptively aligns the sub-motion features to the text embeddings in the CLIP space in an MoE fashion. The extensive experiments demonstrate that OMG achieves state-of-the-art zero-shot performance on text-to-motion generation.
We believe it is a significant step towards open-vocabulary motion generation of human characters, with wide potential applications in movies, games, robotics, and VR/AR.


\section*{Acknowledgement}
This work was supported by National Key R\&D Program of China (2022YFF0902301), Shanghai Local college capacity building program (22010502800). We also acknowledge support from Shanghai Frontiers Science Center of Human-centered Artificial Intelligence (ShangHAI).

{
    \small
    \bibliographystyle{ieeenat_fullname}
    \bibliography{main}
}

\clearpage
\section*{\hfil {\LARGE Appendix}\hfil}


\renewcommand\thesection{\Alph{section}}
\renewcommand*{\theHsection}{appedix.\thesection}
\setcounter{section}{0}

\renewcommand{\thefigure}{\Alph{figure}}
\setcounter{figure}{0}
\renewcommand{\thetable}{\Alph{table}}
\setcounter{table}{0}

\appendix

This appendix provides more qualitative results (\cref{sec:appendix:qualitative}), dataset details (\cref{sec:appendix:dataset}), 
user study (\cref{sec:appendix:userstudy}), 
and limitations (\cref{sec:appendix:limitations}).\\
\vspace{-8pt}

\section{Qualitative Results}
\label{sec:appendix:qualitative}

\begin{figure*}[h]
    \centering
    \includegraphics[width=1.05\linewidth]{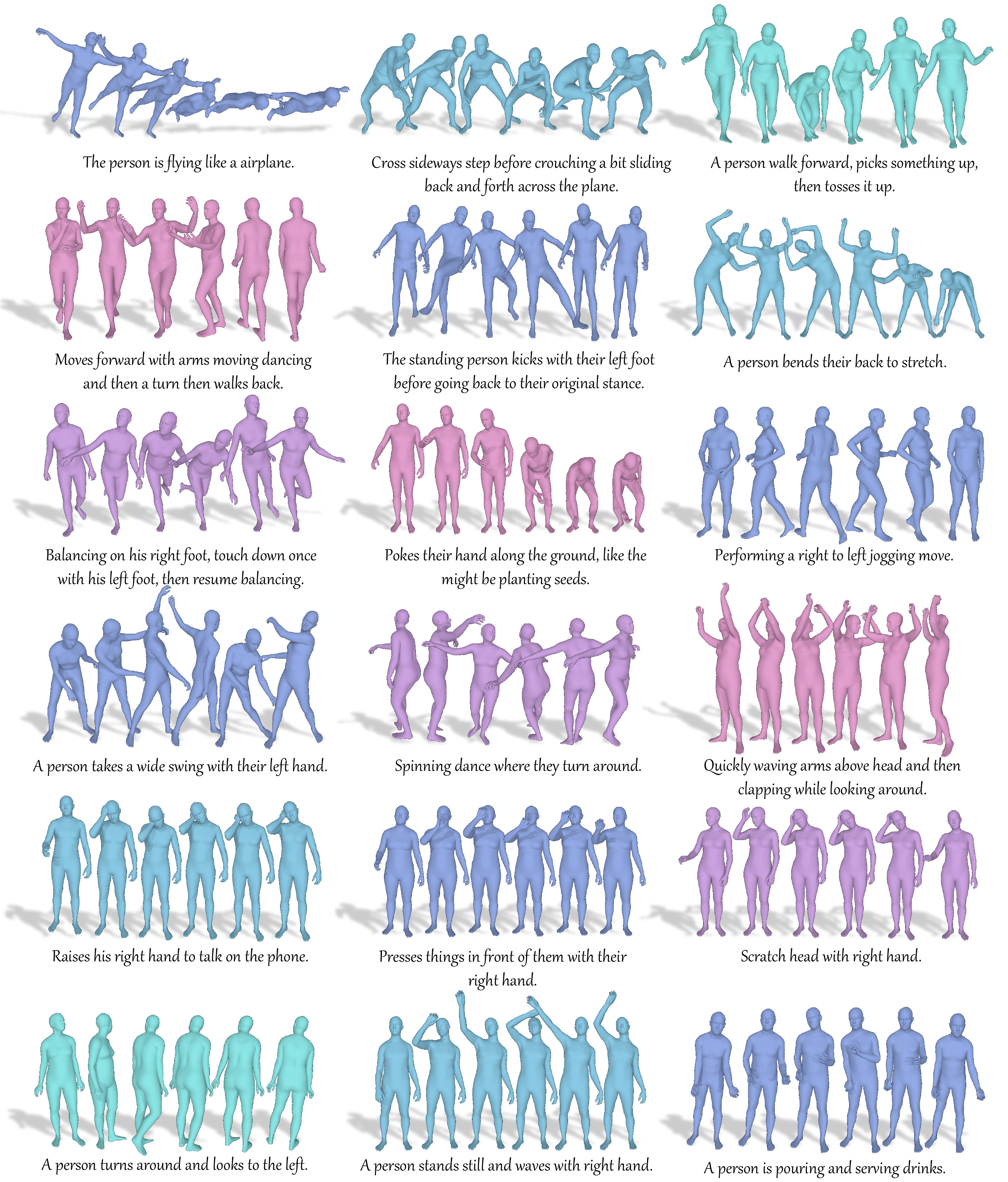}
    \caption{Qualitative results on HumanML3D test set.}
    \label{fig:appendix:humanml}
\end{figure*}

\begin{figure*}[h]
    \centering
    \includegraphics[width=1.05\linewidth]{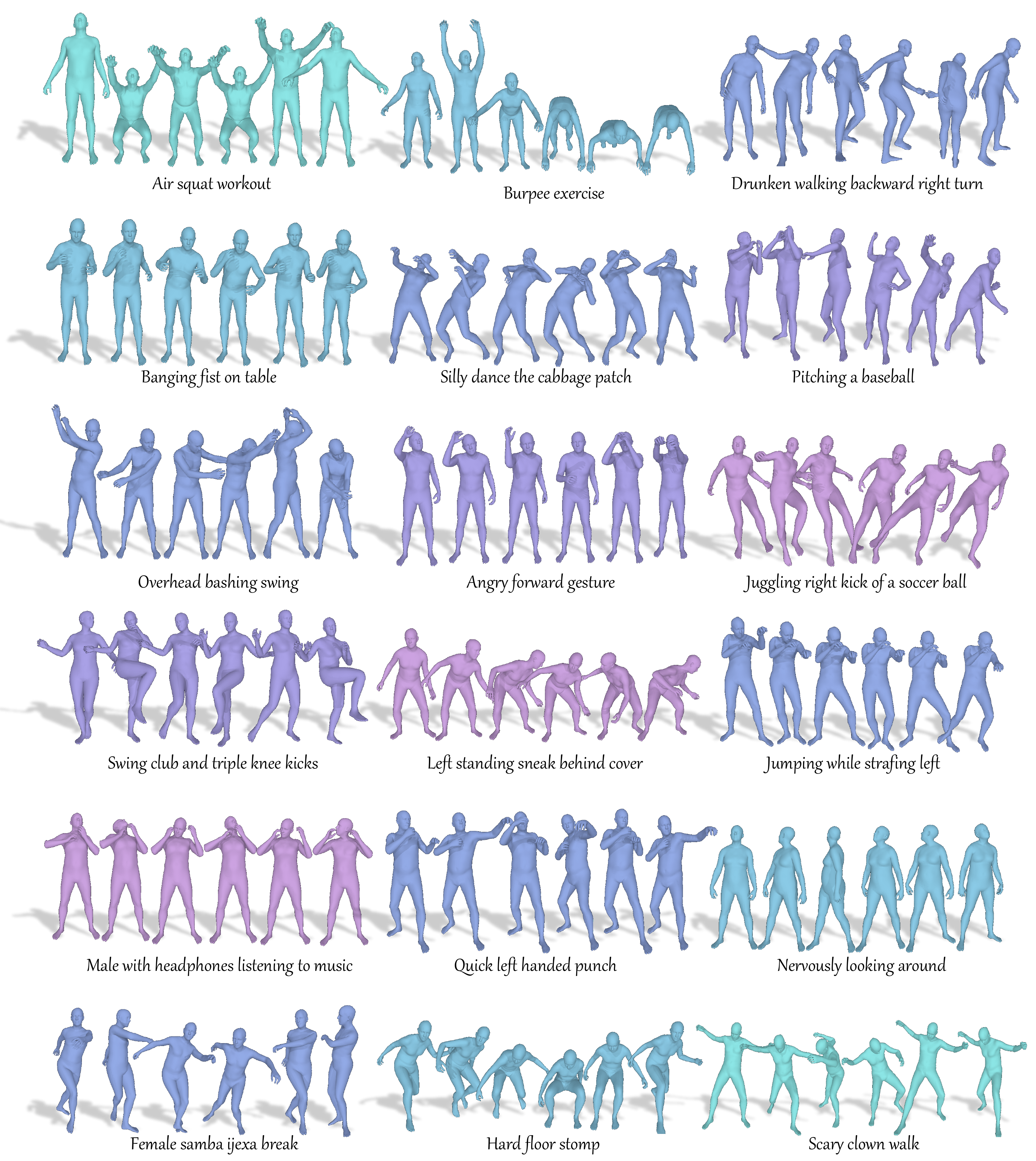}
    \caption{Qualitative results on Mixamo test set.}
    \label{fig:appendix:mixamo}
\end{figure*}

\begin{figure*}[h]
    \centering
    \includegraphics[width=1.05\linewidth]{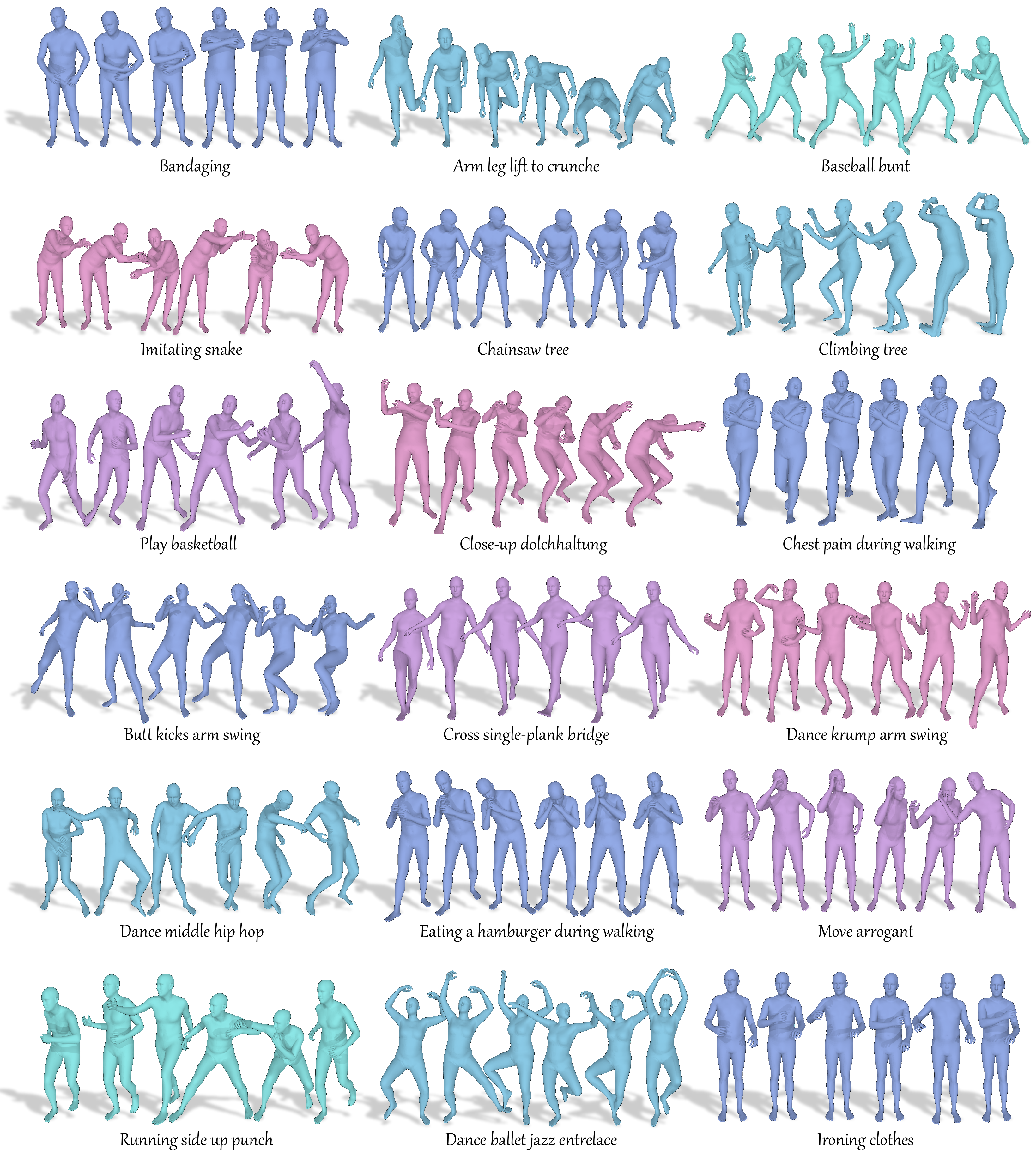}
    \caption{Qualitative results on Motion-X dataset.}
    \label{fig:appendix:motionx}
\end{figure*}

We show more qualitative results from both in-domain and out-of-domain text inputs of several text-motion datasets.
First, we show the generated motions of our OMG model from the text inputs in the HumanML3D test set. 
As illustrated in \cref{fig:appendix:humanml}, our model enables realistic and diverse motion generation from complicated natural sentences.
Then, we show the out-of-domain generation capability using text inputs of the Mixamo test set and the concurrent Motion-X~\cite{lin2023motionx} dataset.
As illustrated in \cref{fig:appendix:mixamo} and \cref{fig:appendix:motionx}, our model well-handles unseen high-level descriptions of motion traits, like ``scary clown'' or ``imitating snake''.

\section{Dataset Details}
\label{sec:appendix:dataset}
Here we provide the details of the motion-only datasets used at the pre-training stage, as illustrated in \cref{fig:datasets1}.
We utilize 13 publicly available human motion datasets captured from various motion modalities, such as artist-created datasets~\cite{harvey2020robust,mason2022local}, marker-based ~\cite{mahmood2019amass,taheri2020grab,araujo2023circle,liu2022beat,ionescu2013human3}, IMU-based~\cite{liang2023hybridcap,trumble2017total} and multi-view markerless~\cite{zhang2022egobody, li2021ai, cai2022humman, liang2023intergen} motion capture datasets, totaling over 22 million frames.  
Since the majority of motion data is in SMPL format, we apply the retargeting algorithm to standardize them to the SMPL skeleton with rotations and positions of 22 joints, and global translation.

Moreover, we utilize HumanML3D~\cite{guo2022generating} training set to train our motion ControlNet for fair comparisons with previous methods. 
The dataset consists of 14616 motion clips with 44970 text annotations, totaling 3.1M motion instances, as illustated in \cref{fig:datasets2}.
We further introduce Mixamo~\cite{Mixamo} dataset, consisting of abundant artist-created animations and human-annotated descriptions.
It is widely used in character animation applications, such as games and VR/AR.
We employ it to benchmark the zero-shot performance due to its wide variety of diverse and dynamic motions and complicated and abstract motion trait descriptions.

\begin{table}[tbp]
	\centering
    \resizebox{1\linewidth}{!}{
    
        \begin{tabular}{@{}lcccc@{}}
        
        \toprule
        Dataset & Duration (h) & Frame Number & Mocap Modality & Motion Format \\
        \midrule
        HCM~\cite{liang2023hybridcap} & 2.9 & 0.3M & IMU & SMPL  \\
        AMASS~\cite{mahmood2019amass} & 62.9 & 6.8M & Marker & SMPL \\
        EgoBody~\cite{zhang2022egobody} & 0.4 & 0.04M & RGB-D & SMPL \\
        GRAB~\cite{taheri2020grab} & 3.8 & 0.4M & Marker & SMPL  \\
        AIST++~\cite{li2021ai} & 4.0 & 0.4M & RGB & SMPL  \\
        HuMMan~\cite{cai2022humman} & 0.9 & 0.1M & RGB-D & SMPL \\
        InterHuman~\cite{liang2023intergen} & 13.1 & 1.4M & RGB & SMPL   \\
        CIRCLE~\cite{araujo2023circle} & 10.0 & 1.1M & Marker & SMPL   \\
        BEAT~\cite{liu2022beat} & 76 & 8.2M & Marker & BVH   \\
        LaFan1~\cite{harvey2020robust} & 4.6 & 0.5M & Marker & BVH   \\
        Human3.6M~\cite{ionescu2013human3} & 5.0 & 0.5M & Marker & SMPL   \\
        Total Capture~\cite{trumble2017total} & 0.8 & 0.09M & IMU & SMPL   \\
        100style~\cite{mason2022local} & 22.1 & 2.4M & marker & BVH   \\
        
        \midrule
        Total & 206.5 & 22.3M & - & -   \\
        \bottomrule
        \end{tabular}
    }
    \caption{The details of unlabeled motion datasets used at the pre-training stage.}
    \label{fig:datasets1}
\end{table}

\begin{table}[tbp]
	\centering
    \resizebox{1\linewidth}{!}{
    
        \begin{tabular}{@{}lcccccc@{}}
        
        \toprule
        Dataset & Clip Number & Text Number & Duration (h) & Frame Number  & Motion Format \\
        \midrule
        HumanML3D~\cite{Guo_2022_CVPR} & 14616 & 44970 & 28.59 & 3.1M & SMPL  \\
        Mixamo~\cite{Mixamo} & 2254 & 2254 & 2.5 & 0.3M  & FBX  \\
        \bottomrule
        \end{tabular}
    }
    \caption{We use HumanML3D training set at the fine-tuning stage and HumanML3D and Mixamo test set for evaluation.}
    \label{fig:datasets2}
\end{table}

\begin{figure}[tbp]
    \centering
    \includegraphics[width=1\linewidth]{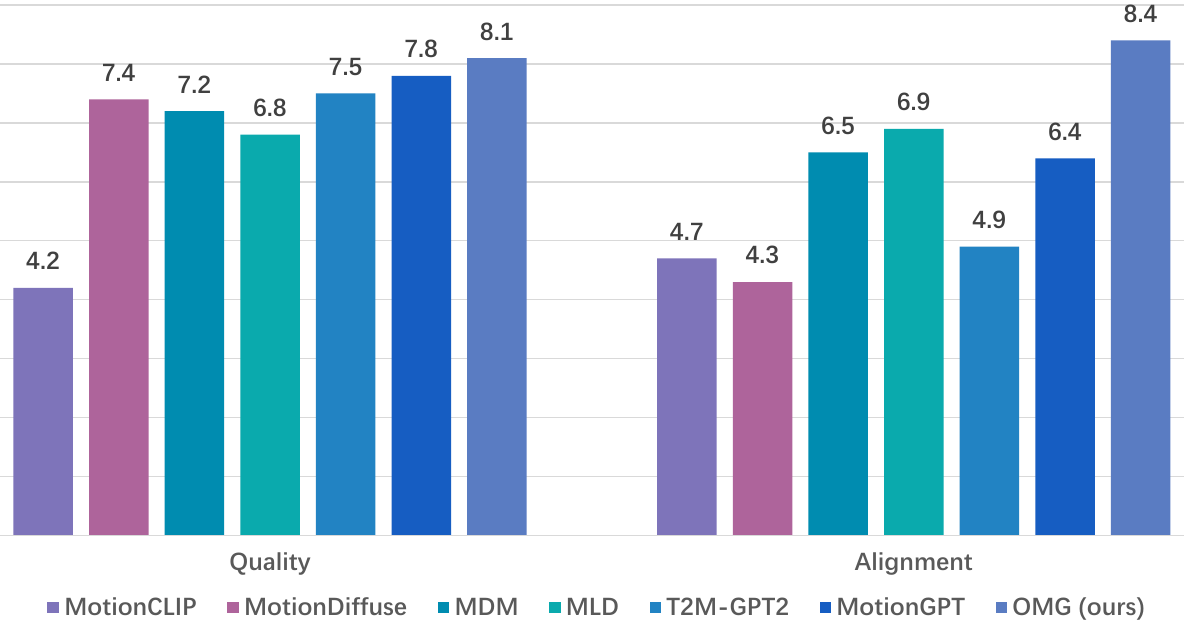}
    \caption{\textbf{User Study.} We show the average quality rates and the average alignment rates of the compared methods, which indicate human evaluation of both motion quality and text-motion consistency respectively. }
    \label{fig:appendix:user}
\end{figure}
\vspace{-8pt}

\section{User Study}
\label{sec:appendix:userstudy}
\begin{figure*}[tbp]
    \centering
    \includegraphics[width=0.9\linewidth]{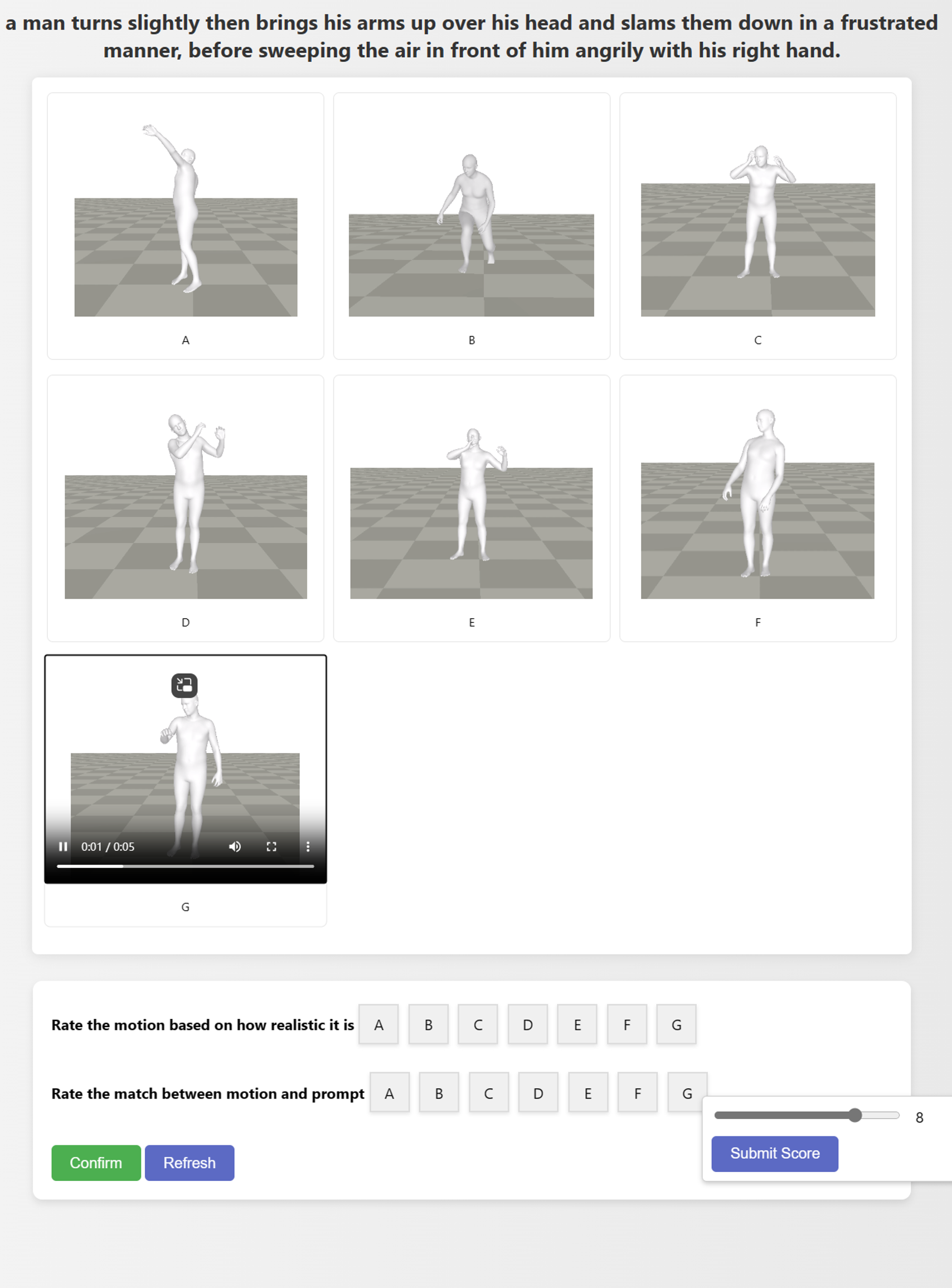}
    \caption{\textbf{User Study.} We ask 20 users to rate the motion quality and text-motion consistency of 60 results generated from each method. The rating range is from 1 to 10.}
    \label{fig:appendix:interface}
\end{figure*}

For the comparisons of the user study presented in \cref{fig:appendix:user}, we ask the users to ``Rate the motion based on how realistic it is'' and ``Rate the match between motion and prompt''. 
The provided motions are generated from 60 text descriptions, 30 of which are randomly generated from the HumanML3D~\cite{guo2022generating} test set and 30 from Mixamo~\cite{Mixamo} test set. 
We invite 20 users, shuffle the order of results from the distinct compared methods, and ask them to complete the rating, as illustrated in \cref{fig:appendix:interface}. 
As shown in~\cref{fig:appendix:user}, our OMG was preferred over the other state-of-the-art methods in both motion quality and text-motion alignment.

\section{Limitations}
\label{sec:appendix:limitations}
As the trial to explore realistic open-vocabulary motion generation, the proposed OMG still has limitations as follows.

\myparagraph{Motion space.}
Our method still relies on the training motion manifold and cannot generate motions that are beyond the scope of the training data, such as flying, yoga, or swimming. 

\myparagraph{Precise control.}
Our method does not explicitly model the temporal order and inclusion relations of sub-motions, which are unable to handle precise control, such as picking an object or reaching a goal.

\myparagraph{Physically implausible.}
Our method does not explicitly model physical dynamics, which leads to physically implausible motion generation.
Recent physics-based motion control~\cite{yuan2023physdiff, InterPhysHassan2023,yao2023moconvq} approaches use reinforcement learning to control human characters in a physically simulated environment, achieving impressive motion quality.
It’s interesting to introduce physics into the conditional generative model pipeline.

\myparagraph{Maximum length.}
Same as most motion generation methods, our method can generate arbitrary
length results but still under the max-length in the dataset.
It’s interesting to model a non-stop human motion in temporal consistency.

\myparagraph{Full-body dynamics.}
Our method focuses on articulated human bodies.
How to model the full-body dynamics including the face, eyes, hands, and even toes, which enables complicated interactions with our complex physical world, remains a huge challenge.

\end{document}